\documentclass{article}

\usepackage[utf8]{inputenc} 
\usepackage[T1]{fontenc}    
\usepackage[preprint,nonatbib]{arxiv}
\usepackage{graphicx} 
\usepackage[utf8]{inputenc}
\usepackage[hidelinks]{hyperref}

\usepackage{subcaption}
\usepackage[]{natbib}
\setcitestyle{square}

\usepackage{etoolbox}
\makeatletter
\renewcommand{\bibinfo}[2]{%
	\@ifstrequal{#1}{doi}{}{#2}%
}
\makeatother

\title{\LARGE \bf
Acquisition of high-quality images for camera calibration in robotics applications\\ via speech prompts}

\usepackage{authblk}

\newbox{\orcid}\sbox{\orcid}{\includegraphics[scale=0.06]{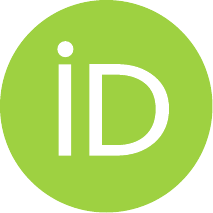}}
\author[ 1]{%
	\href{https://orcid.org/0000-0001-8532-0262}{\usebox{\orcid}\hspace{1mm}Timm Linder\thanks{\texttt{first.last@de.bosch.com}} }%
}
\author[2]{%
	\href{https://orcid.org/0009-0000-2239-8230}{\usebox{\orcid}\hspace{1mm}Kadir Yilmaz\thanks{\texttt{lastname@vision.rwth-aachen.de}}\hspace{1mm}}}%

\author[1]{David Adrian}%

\author[2]{%
	\href{https://orcid.org/0000-0003-4225-0051}{\usebox{\orcid}\hspace{1mm}Bastian Leibe}}%

\affil[1]{Bosch Corporate Research \& Bosch Center for AI, Renningen, Germany}
\affil[2]{Computer Vision Group, RWTH Aachen University, Germany}

\renewcommand{\headeright}{}
\renewcommand{\undertitle}{Preprint}
\renewcommand{\shorttitle}{Camera calibration via speech prompts}

\hypersetup{
	pdftitle={Acquisition of high-quality images for camera calibration in robotics applications via speech prompts},
	pdfsubject={cs.RO, cs.CV},
	pdfauthor={Timm Linder},
	pdfkeywords={Camera calibration, robotics, speech, transcription, motion blur},
}

\date{}
\begin{document}

\maketitle

\begin{abstract}
Accurate intrinsic and extrinsic camera calibration can be an important prerequisite for robotic applications that rely on vision as input.
While there is ongoing research on enabling camera calibration using natural images, many systems in practice still rely on using designated calibration targets with e.\,g. checkerboard patterns or April tag grids.
Once calibration images from different perspectives have been acquired and feature descriptors detected, those are typically used in an optimization process to minimize the geometric reprojection error.
For this optimization to converge, input images need to be of sufficient quality and particularly sharpness; they should neither contain motion blur nor rolling-shutter artifacts that can arise when the calibration board was not static during image capture.
In this work, we present a novel calibration image acquisition technique controlled via voice commands recorded with a clip-on microphone, that can be more robust and user-friendly than e.\,g. triggering capture with a remote control, or filtering out blurry frames from a video sequence in postprocessing. To achieve this, we use a state-of-the-art speech-to-text transcription model with accurate per-word timestamping to capture trigger words with precise temporal alignment. Our experiments show that the proposed method improves user experience by being fast and efficient, allowing us to successfully calibrate complex multi-camera setups.
\end{abstract}


\section{INTRODUCTION}

\begin{figure}
	\centering
	\includegraphics[width=\columnwidth]{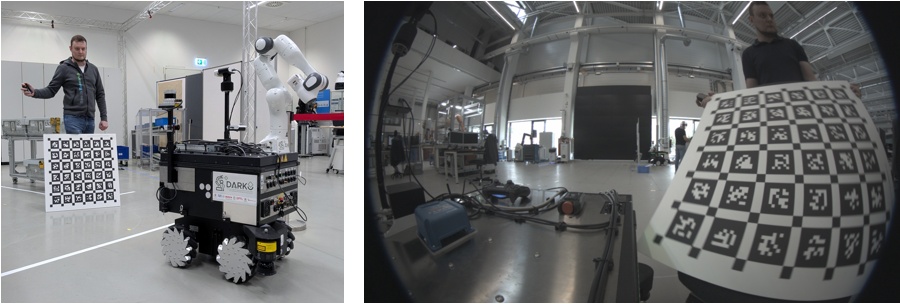}
	\caption{Our use-case is the calibration of pinhole and fisheye cameras in robotics applications using dedicated calibration targets (e.\,g. checkerboard patterns, April tag grids).
	}
	\label{fig:robot-use-case-1}
	\vspace*{0mm}
\end{figure}

Many applications in 2D and 3D perception for robotic systems rely on accurate camera calibration, for instance methods for object detection \citep{Detection:Survey3D:Liang,Detection:Survey6DoF:Guan}, simultaneous localization and mapping (SLAM) \citep{SLAM:Survey:Damjanovic}, or robotic manipulation \citep{Detection:PoseEstManipulation:An}. One concrete example is the monocular 3D human pose estimation method MeTRAbs \citep{HPE:Metrabs:Sarandi}, that we utilize on our mobile robotics platform (Figure~\ref{fig:robot-use-case-1}) for human-robot interaction tasks and human-aware navigation \citep{Navigation:HumanAwareMPC:Stefanini} in agile production and intralogistics scenarios. The approach requires knowledge of the pinhole camera intrinsics as input in order to perform absolute pose recovery. Likewise, its recently proposed extension to fisheye optics \citep{HPE:FisheyeMetrabs:Kaes} requires calibration with a fisheye camera model. 

\begin{figure}
	\centering
	\includegraphics[width=\columnwidth]{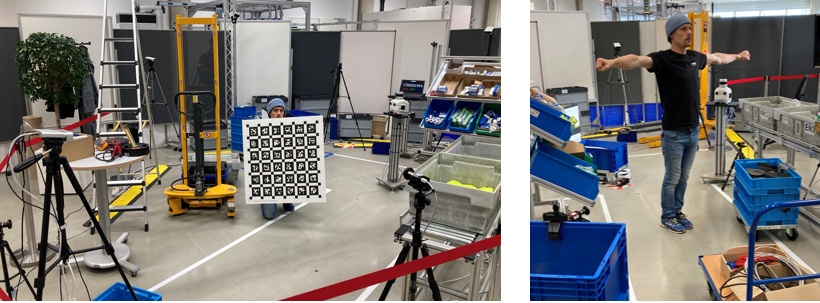}
	\caption{The proposed speech-guided image acquisition method can also be used to calibrate complex multi-view systems, as in the example setup here with >10 cameras on tripods to obtain groundtruth 3D human poses via multi-view triangulation.
	}
	\label{fig:robot-use-case-2}
	\vspace*{0mm}
\end{figure}

Furthermore, to record training data e.\,g. for 3D human pose estimation approaches, complex multi-view camera setups are typically used (Figure~\ref{fig:robot-use-case-2}), often containing around a dozen cameras that are spread out through the scene. In addition to the camera intrinsics, also extrinsics are required in this case, in order to be able to triangulate accurate groundtruth 3D human poses.

The same applies when calibrating multiple cameras on a robot relative to each other, for example to achieve 360-degree surround view coverage on an omnidirectional drive robot like ours, for downstream tasks such as multi-modal 3D object detection \citep{ObjectDetection:RGBDCubeRCNN:Piekenbrinck}, or multi-modal human detection using cameras and lidar \citep{HumanDetection:CrossModalAnalysis:Linder}.

While ongoing research explores on how to perform calibration using in-the-wild images \citep{Calibration:PerspectiveFields:Jin}, in our work we assume that a calibration target (April tag grid or checkerboard pattern) is used for camera calibration, by placing it in different poses relative to the camera. Based upon our extensive experience with calibrating the aforementioned multi-camera systems, using both open-source \citep{Calibration:Kalibr:Website,Calibration:Kalibr:Paper} and commercially available \citep{Calibration:CalibIO:Website} calibration software, we observe that it is crucial to acquire high-quality calibration images to obtain good results. To achieve this goal, we propose a novel technique for speech-guided calibration image acquisition, which allows the operator to securely hold the calibration target with both hands, without requiring any additional support person during the calibration process.

\section{METHOD}

\paragraph{\textbf{Problem definition}} The main challenge that we aim to address with our method is the elimination, or reduction, of calibration images that exhibit motion blur and rolling shutter artifacts, which can cause feature extraction and subsequently camera calibration to fail. In our experience, this issue occurs frequently when recording continuous calibration video sequences while the operator is in motion along with the calibration pattern, as illustrated in the example of Figure~\ref{fig:blur-detection}. 

We also observe that motion blur can become more severe under difficult lighting situations that are often encountered in industrial factory and warehouse environments (Figure~\ref{fig:robot-use-case-1}, right).

\subsection{Alternative approaches}

\begin{figure}[t]
	\centering
	\captionsetup[subfigure]{justification=centering}
	\hspace*{-1.0em}
	\begin{subfigure}{0.3\textwidth}
		\includegraphics[height=2.5cm]{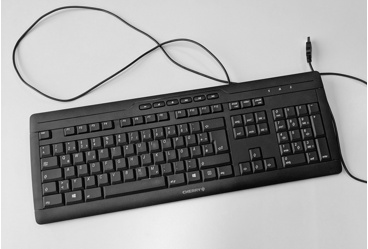}%
		\caption{Wired keyboard or\newline mouse as trigger}
		\label{fig:keyboard}%
	\end{subfigure}
	\hspace{0.2em}
	\begin{subfigure}{0.3\textwidth}
		\includegraphics[height=2.5cm]{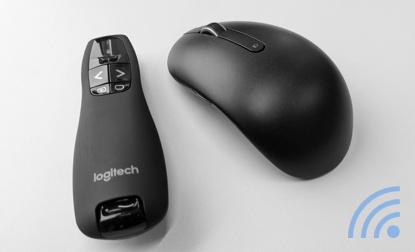}%
		\captionsetup{margin={1em,0em}}
		\caption{Wireless remotes\newline\hspace*{-2em}e.g. Bluetooth presenter}
		\label{fig:presenter}%
	\end{subfigure}
	\hspace{1.3em}
	\begin{subfigure}{0.3\textwidth}
		\includegraphics[height=2.5cm]{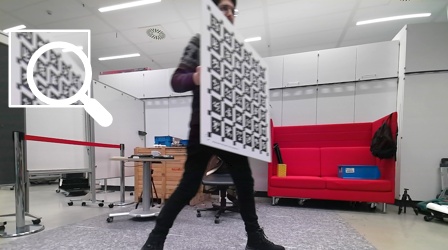}%
		\captionsetup{margin={2em,0em}}
		\caption{Automatic detection\newline\hspace*{-1em}of blurry frames/motion}
		\label{fig:blur-detection}%
	\end{subfigure}
	\caption{Alternative approaches that we evaluated, with different shortcomings that make them impractical when calibrating complex multi-camera systems spread out over larger areas.}
	\label{fig:alternative-approaches}
\end{figure}

Before introducing the proposed method, we briefly review other related approaches that we have experimented with, and that did not yield satisfactory results.

\paragraph{\textbf{Adjusting acquisition parameters}} By reducing exposure time, increasing sensor gain, improving lighting, or switching to a more light-sensitive camera/lens combination, the amount of motion blur can be reduced. However, not all of these parameters are always under the user's control. For example, in our case, we are also interested in experimenting with very low-cost sensors and lenses, that start to exhibit motion blur (or heavy pixel noise at high gain) already under benign lighting conditions during daylight.

\paragraph{\textbf{Wired triggering}} Using a wired keyboard or mouse as input (Figure~\ref{fig:keyboard}) requires the operator to put the calibration target to rest, e.\,g. using a tripod, and then walk to the computer to press a key in ordder to trigger image acquisition. As this process can be very time-consuming, it is not a practical option in multi-camera setups that extend over a larger space, unless a second person assists with the calibration effort, which increases operation cost.

\paragraph{\textbf{Remote-controlled triggering}} During initial experiments with our multi-view capture setup, we therefore tried out several remote-controlled, bluetooth-based triggers (e.\,g. a wireless presenter, see Figure~\ref{fig:presenter}). However, these proved to be unreliable at distances over 3-4m, or when the line-of-sight to the receiver was obstructed by environmental structures or the calibration target. Also, triggering delays sometimes appeared non-deterministic, reaching up to 2-3 seconds in some cases. In our experiments, this often led to the situation that the operator had already started moving again, while the image had not yet been recorded. Furthermore, the operator may need both hands to securely hold the calibration target, thus hitting keys on a remote while not introducing additional motion jitter can be an intricate endeavour.

\paragraph{\textbf{Automated detection of low-quality calibration images}} Various metrics for detecting and quantifying blurriness of images with (motion) blur artifacts (Figure~\ref{fig:blur-detection}) have been proposed \citep{Blur:Survey:Andrade}, e.\,g. based upon variance of the Laplacian, other handcrafted metrics \citep{ImageQuality:BlurDetection:Liu}, or deep neural networks \citep{ImageQuality:MotionBlurDetection:Beomseok}. The more general task is called blind image quality assessment \citep{ImageQuality:ReIQA:Saha}, if no groundtruth image is given. The challenge here is to find an approach that generalizes well over different camera/lens combinations and scene compositions (foreground, background objects), without having to adjust manually thresholds for every scene or frame.

\paragraph{\textbf{Motion detection}} Finally, one could also attempt to curate images from a video where the calibration target is static, by detecting motion across a neighboring set of frames using a sliding-window based technique, e.\,g. using optical flow \citep{OpticalFlow:Survey:Alfarano}. However, this easily fails if there are objects in the scene that naturally exhibit motion, e.\,g. ceiling fans, moving people in the background, or the constantly blinking LED strips that we use for intention communication on our robot.

\subsection{Our proposed method}

\paragraph{\textbf{Overview}} We propose to utilize speech commands to trigger image acquisition whenever the operator has positioned the calibration target in a new steady pose (for example by placing it on a stand, chair, or a forklift in our case). We use a configurable trigger word like ``Capture!'' for this purpose. By using speech as opposed to a remote control, the operator has both hands free to move around or securely hold the calibration target. For full control over the image extraction process, we record aligned, continuous video and audio sequences, and then extract candidate frames based upon the speech prompts offline in post-processing. For multi-camera setups where also extrinsics need to be computed, we ensure to perform image extraction in a synchronized fashion, based on ROS message header timestamps.

\paragraph{\textbf{Audio-video synchronization}} To align separately recorded audio from a clip-on microphone with the video recordings from one or multiple cameras, the operator marks the beginning of a new calibration sequence by clapping the hands, which is clearly discernible in both audio and video. While we currently manually determine the claps timestamps, which is required only once per session, this could also be automated using e.\,g. hand/body pose estimation and audio signal detection.

\begin{figure}
	\centering
	\includegraphics[width=10cm]{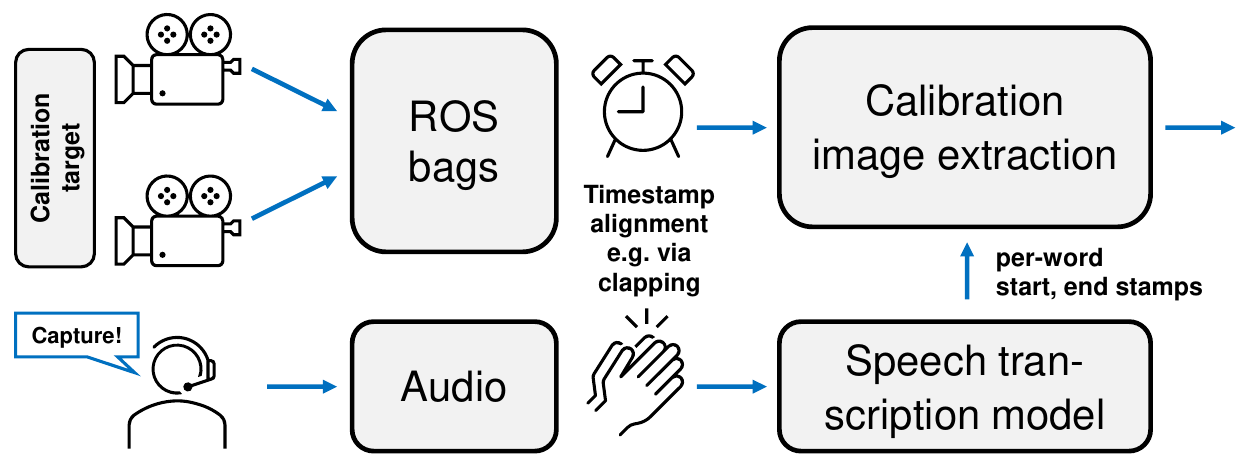}
	\caption{Flow diagram of the proposed approach. In our implementation, we use WhisperX \citep{SpeechRecognition:WhisperX:Bain} as the speech transcription model, which provides highly accurate per-word timestamps.}
	\label{fig:flow-diagram}
	\vspace*{0mm}
\end{figure}

\paragraph{\textbf{Detection of trigger words}} To detect and temporally localize the trigger words, we use a speech transcription model based upon Whisper \citep{SpeechRecognition:Whisper:OpenAI}, a state-of-the-art general-purpose speech recognition model trained using weak supervision on a large-scale multilingual audio dataset ($\sim$680,000 hours). As a multitasking model, it can perform speech recognition, speech translation, and language identification, however, its per-word timestamps for longer audio sequences are not very accurate, and in our experiments it was sometimes prone to hallucinations. Therefore, we utilize the more recent WhisperX \citep{SpeechRecognition:WhisperX:Bain}, which extends Whisper with accurate word-level timestamping through forced phoneme alignment via wav2vec2 \citep{SpeechRecognition:Wav2Vec2:Baevski} as well as voice activity detection to reduce hallucinations, both of which proved to be very helpful in our scenario. We use the mean of the start and end timestamps of the detected trigger word as timestamp for image extraction.

\paragraph{\textbf{Further processing}} As shown in Fig.~\ref{fig:flow-diagram}, once the timestamps of all ``Capture!'' commands have been determined using the proposed method, we extract the corresponding images and feed them into a commercially available camera calibration suite \citep{Calibration:CalibIO:Website}, to obtain intrinsics and extrinsics for our pinhole and fisheye cameras. 

\begin{figure}
	\centering
	\includegraphics[width=\columnwidth]{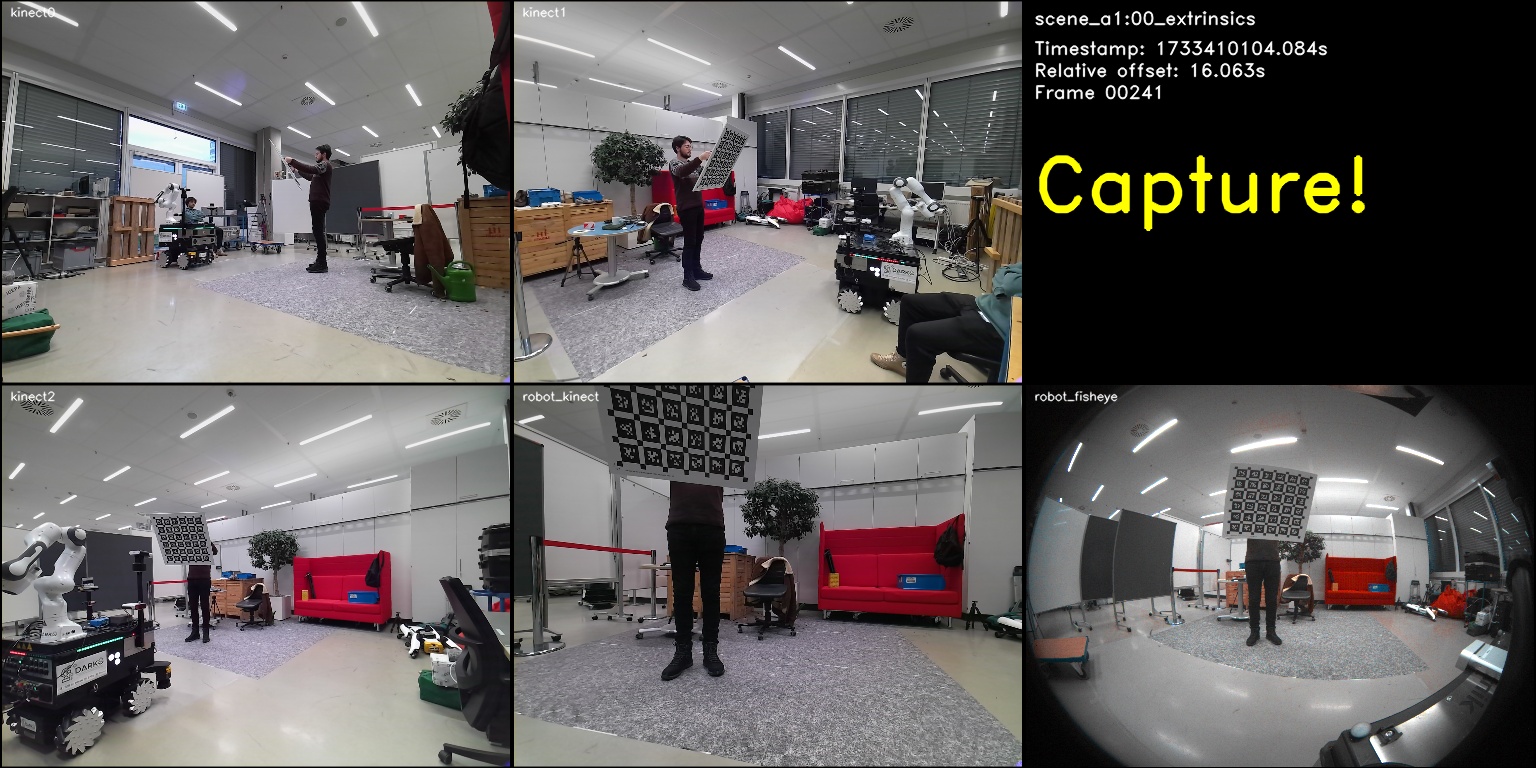}
	\caption{Our user interface shows an operator recording calibration images in a synchronized multi-camera setup involving our robot platform (lower left). He prompts image acquisition during steady poses via speech commands, while having his hands free to securely hold and move the calibration target. The recognized trigger word is shown on the right in yellow.
	}
	\label{fig:multi-camera-setup-with-robot}
	\vspace*{0mm}
\end{figure}

\paragraph{\textbf{User interface}} To visualize the synchronized image frames from our multi-view camera setups, along with the extracted trigger words, we have further implemented a graphical user interface using OpenCV, shown in Figure~\ref{fig:multi-camera-setup-with-robot}, helping us verify the precise alignment of extracted trigger timestamps.

\section{EXPERIMENTS}

We now want to demonstrate that the proposed calibration image acquisition technique can be successfully used to calibrate cameras on a robotic system.

\paragraph{\textbf{Experimental setup}}  For our experiments, we recorded calibration video sequences of around 5 minutes length using 4 Azure Kinect pinhole RGB cameras and one 5.1MP machine vision fisheye camera as ROS bagfiles. One Kinect and the fisheye camera were mounted on our robot and connected to its internal PC, while three of the Kinects have been placed on tripods and connected to an external computer that records also the audio from a low-cost clip-on microphone via a USB radio receiver. Both computers' clocks have been synchronized via network time protocol (NTP), the clock of the machine vision camera is synchronized via precision time protocol (PTP). Around 50 speech prompts for image acquisition have been recorded per session and were stored as Ogg Vorbis audio files. We use an April tag grid with 6 rows and 6 columns, printed on a 1x1 meter rigid surface to which we attached handholds using aluminium profiles, as calibration target.

\paragraph{\textbf{Camera models for calibration}} For calibration of pinhole images, we use the standard OpenCV pinhole camera model with 3 radial and 2 tangential distortion coefficients. For calibration of fisheye images, we use the Double-Sphere model by \cite{Calibration:DoubleSphere:Usenko}, which yielded robust calibration results on 5 different models of fisheye lenses that we experimented with.

\paragraph{\textbf{Qualitative results}} To understand whether our method helps in obtaining high-quality calibration images, we visually inspect the extracted image frames and manually assess them for their quality (sharpness, lack of motion blur). Thanks to our proposed approach, no noticeable motion blur has been observed in any image, as the operator consistently issued speech prompts while he was standing still, such that the calibration target was resting in a stationary pose.

\begin{figure}
	\centering
	\includegraphics[width=1.0\columnwidth]{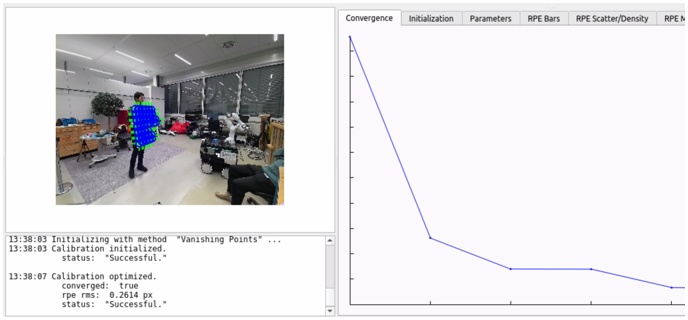}
	\caption{Successfully converging calibration using an existing calibration software \citep{Calibration:CalibIO:Website} and motion blur-free calibration images extracted using our proposed approach.}
	\label{fig:calib-result}
	\vspace*{0mm}
\end{figure}

\paragraph{\textbf{Quantitative evaluation}} We then use the extracted calibration images to perform actual calibrations using existing commercial calibration software \citep{Calibration:CalibIO:Website}. As visualized exemplarily in Figure~\ref{fig:calib-result}, the calibration using the images extracted via speech prompt quickly converges. We obtain root mean square reprojection errors of on average $<$0.5\,px, which can be considered a successful calibration result.

\paragraph{\textbf{Video material}} Our supplementary video\footnote{\url{https://www.youtube.com/watch?v=HFjTqGHMxIw}} provides further examples and insights into our proposed calibration image acquisition process. At the end, the video also includes a working demonstration of our user interface from Figure~\ref{fig:multi-camera-setup-with-robot}, showing trigger words and associated timestamps as they are being extracted by the proposed approach.

\paragraph{\textbf{Study on user satisfaction}} In a small user study, three different subjects that we interviewed noted that the proposed method for calibration image acquisition is more convenient to use than 1) a bluetooth-based trigger, 2) running back and forth between calibration board and PC to trigger frame capture, 3) performing calibration on the entire video sequence. It is noteworthy that the latter can take around 6 hours due to the lengthy feature detection process with such a large number of frames and multiple cameras. Therefore, prior extraction of relevant key frames is essential, which is greatly simplified by our proposed technique.

\section{CONCLUSIONS}

In this paper, we proposed a novel technique for acquisition of high-quality calibration images via speech prompts. Our method allows the operator to securely hold the calibration target still with both hands, without requiring any additional support person during the calibration process, making the process robust, fast and efficient.

Using the proposed technique, which leverages a state-of-the-art AI-based method for accurately timestamped speech recognition, we have been easily able to calibrate multiple multi-camera setups. The calibrated camera setups have been used to record novel datasets for our ongoing research. We believe that our method and lessons learned can help robotics researchers to streamline their camera calibration workflows.

In future work, we want to explore how we can utilize the proposed setup also for speech-guided 3D scene labelling tasks, where precise per-word timestamps are of equally high importance as semantically accurate speech-to-text transcription quality, and powerful multi-task language models such as WhisperX can demonstrate their full potential.

\acksection{The preparation of this manuscript and its experiments have been supported by the EU Horizon 2020 research and innovation program under grant agreement 101017274 (DARKO). Support in data acquisition and ideation was provided by the project ``Context Understanding for Autonomous Systems'' by Robert Bosch GmbH.}

\bibliographystyle{plainnat}

\end{document}